\pdfoutput=1

\documentclass[11pt]{article}

\usepackage[preprint]{acl}

\usepackage{times}
\usepackage{latexsym}

\usepackage[T1]{fontenc}

\usepackage[utf8]{inputenc}

\usepackage{microtype}

\usepackage{inconsolata}

\usepackage{multirow} 
\usepackage{booktabs} 
\usepackage{bbding}
\usepackage{graphicx}
\usepackage{amssymb}
\usepackage{colortbl}
\usepackage{amsmath} 
\usepackage[normalem]{ulem}
\usepackage{inconsolata}
\usepackage{arydshln}
\useunder{\uline}{\ul}{}
\title{Explanation based In-Context Demonstrations Retrieval for Multilingual Grammatical Error Correction}

\author{Wei Li, Wen Luo, Guangyue Peng, Houfeng Wang \\
  State Key Laboratory of Multimedia Information Processing \\
  School of Computer Science, Peking University \\
  weili22@stu.pku.edu.cn, llvvvv22222@gmail.com,\{agy, wanghf\}@pku.edu.cn}

\begin{document}
\maketitle
\begin{abstract}
Grammatical error correction (GEC) aims to correct grammatical, spelling, and semantic errors in natural language text. With the growing of large language models (LLMs), direct text generation has gradually become the focus of the GEC methods, and few-shot in-context learning presents a cost-effective solution. However, selecting effective in-context examples remains challenging, as the similarity between input texts does not necessarily correspond to similar grammatical error patterns. In this paper, we propose a novel retrieval method based on natural language grammatical error explanations (GEE) to address this issue. Our method retrieves suitable few-shot demonstrations by matching the GEE of the test input with that of pre-constructed database samples, where explanations for erroneous samples are generated by LLMs. We conducted multilingual GEC few-shot experiments on both major open-source and closed-source LLMs. Experiments across five languages show that our method outperforms existing semantic and BM25-based retrieval techniques, without requiring additional training or language adaptation. This also suggests that matching error patterns is key to selecting examples. Our code is available at \href{https://github.com/GMago-LeWay/FewShotGEC}{https://github.com/GMago-LeWay/FewShotGEC}.

\end{abstract}

\section{Introduction}

\begin{figure}[t]
    \centering
    \includegraphics[width=1.0\columnwidth]{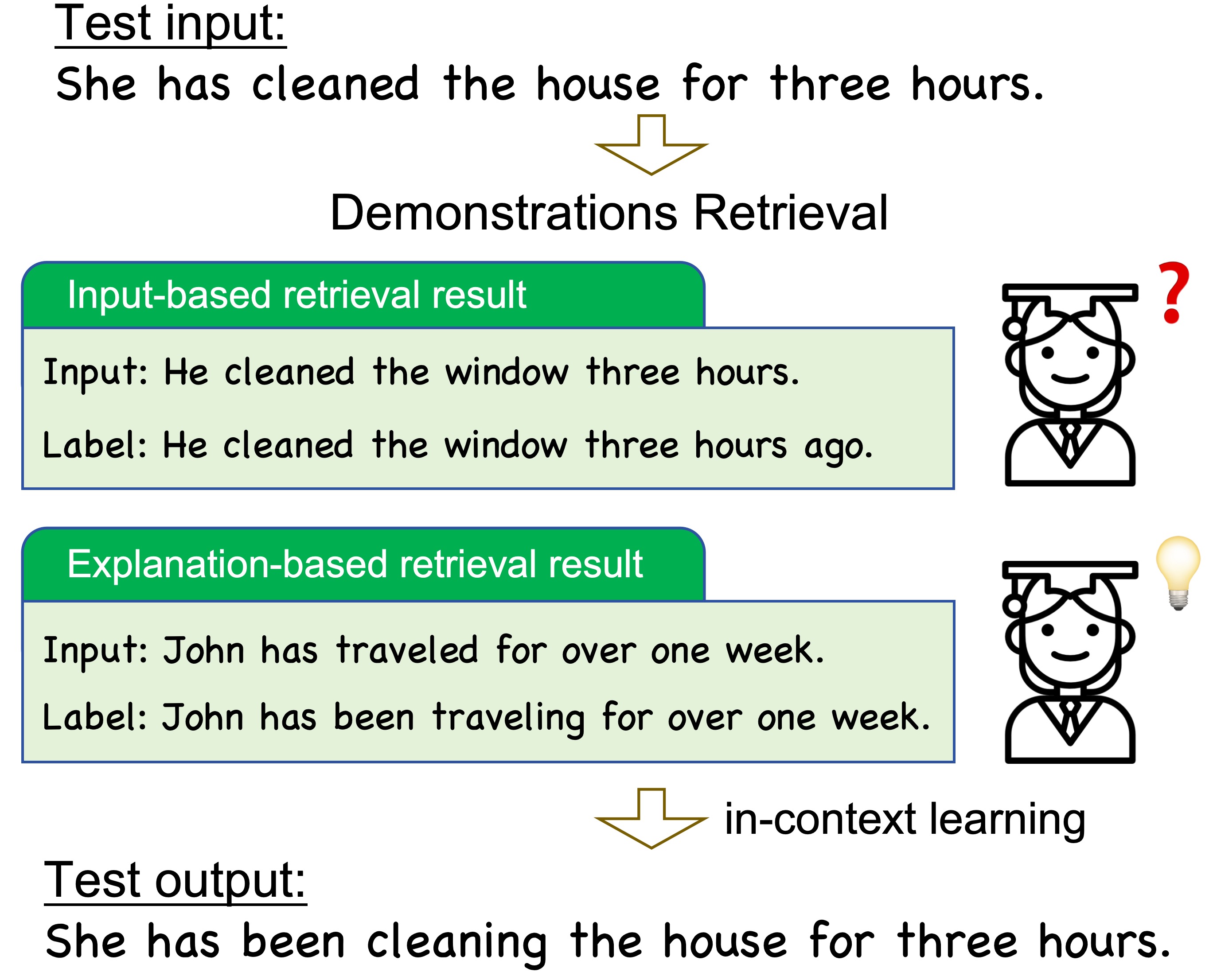}
    \caption{The comparison between input-based demonstrations and explanation-based demonstrations for GEC. Samples with similar inputs do not necessarily contain the same grammatical error patterns. However, through preliminary checks and initial explanations, it is possible to retrieve samples with similar errors from a database indexed by grammatical error explanations, even if the semantics of the demonstrations differs significantly from the test input.}
    \label{fig:intro}
\end{figure}

The goal of grammatical error correction (GEC) is to correct errors in natural language text, including grammatical, spelling, and certain semantic errors \cite{2023gecsurvey}. In the context of language learning, GEC methods can assist learners in correcting mistakes in written text \cite{katinskaia2021assessing_in_learning, caines2023teaching_application_gec}. GEC can be considered a form of machine translation, where the input is a potentially erroneous text, and the output is its corrected version \cite{yuan-briscoe-2016-grammatical}. As a result, text-to-text generation models underpin many GEC approaches \cite{junczys-dowmunt-etal-2018-approaching, katsumata-komachi-2020-stronger}. While edit-based methods exist that modify specific parts of the text for correction \cite{awasthi2019parallel_edit_seq2edit_gec, omelianchuk-etal-2020-gector}, the rise of large language models (LLMs) \cite{brown2020gpt3, touvron2023llama} has made direct text generation the dominant approach in GEC. Given the need for language learning across different languages, multilingual GEC has become an important area of research. This challenge is often tackled by constructing datasets and training models separately for each language \cite{ijcai2022-unified-mgec, rothe2021clang8_gec}. With large-scale pre-training on multilingual data, the performance of LLMs in multilingual contexts is gradually improving \cite{qin2024multilingualllms}, and LLMs are increasingly being applied to the multilingual GEC domain \cite{luhtaru-etal-2024-nllb-gec}.

Few-shot inference can achieve strong performance on downstream tasks through in-context learning (ICL) without the need for fine-tuning LLMs \cite{brown2020gpt3}. Effective ICL depends on selecting appropriate in-context examples, or demonstrations, which can be guided by established principles \cite{agrawal-etal-2023-context}. The selection process can be facilitated through retrieval mechanisms, such as k-nearest neighbors (kNN)-based retrieval \cite{vasselli-watanabe-2023-knngec}. One advantage of example-based GEC systems is that they can enhance system interpretability by providing learners with illustrative examples \cite{kaneko-etal-2022-interpretability}.

For retrieving in-context examples, previous studies primarily use the test input as the query and the input texts of labeled data as keys for the database \cite{retrieve-demonstration-survey}. Matching between query and keys is typically based on semantic or syntactic similarities, or methods like BM25 \cite{hongjin2022selective, tang-etal-2024-ungrammatical, robertson2009probabilistic}. This approach is intuitive since both the test input and the input text of labeled data are from the same domain—if the inputs are similar, the corresponding labels are likely similar as well. However, GEC tasks pose a unique challenge, as grammatical errors are often local and bear little correlation to the overall structure of the text. Consequently, input-based retrieval often fails to retrieve references with analogous grammatical errors.

We propose that a more fundamental solution lies in using the relationship between the input text and the label as the basis for retrieval. Recent works have begun to explore leveraging the interaction between text and label to improve in-context example selection \cite{sun2024retrieved-by-mistake}. For GEC, we hypothesize that this relationship can be effectively captured through grammatical error explanations (GEE) \cite{song-etal-2024-gee}.

In this paper, we introduce a novel retrieval method based on natural language explanations, applicable in multilingual settings. Our method uses GEE as both the query and key for retrieval, aiming to retrieve examples with similar grammatical errors to serve as demonstrations for in-context learning. Without requiring model training, this approach enhances LLM few-shot performance on multilingual GEC tasks. Figure \ref{fig:intro} presents an illustrative example where traditional semantic matching fails to retrieve relevant samples, while explanation-based matching successfully identifies a sample with a similar grammatical error.

To implement this, we first use labeled data and LLMs to generate GEE and construct a database. During inference, the input text undergoes a grammar check, providing an initial explanation, which is then matched against the GEE database to retrieve suitable demonstrations. These demonstrations are used in the ICL process to produce the final grammatical corrections.

We conducted experiments on GEC datasets across five languages. Our proposed method consistently outperforms semantic and BM25 retrieval in terms of the $F_{0.5}$ score across both open-source and closed-source models. The advantages of the proposed method contain: 1) It requires no additional training, leveraging ICL to tap into the intrinsic GEC capabilities of LLMs. Some studies suggest that the $F_{0.5}$ score underestimates LLM GEC performance, with human evaluations yielding better results \cite{coyne2023analyzing}. 2) It is highly extendable to multilingual settings, as the use of natural language explanations ensures applicability across languages. 3) The system is more interpretable, with each demonstration linked to a GEE.

The main contributions of this work are as follows:
\begin{itemize}
    \item We propose a novel demonstration retrieval method for in-context learning of GEC based on natural language explanations and design a GEC process involving grammar checks, retrieval, and final corrections.
    \item We construct GEC databases with grammatical error explanations across multiple languages, which can be applied to different datasets within the same language.
    \item To the best of our knowledge, this work provides the first aligned evaluation of LLMs' few-shot capabilities for GEC, approaching state-of-the-art performance across several languages.
\end{itemize}

\section{Related Works}
\subsection{Grammatical Error Correction}






Methods in the GEC field have been dominated by sequence-to-sequence generation models \cite{junczys-dowmunt-etal-2018-approaching, katsumata-komachi-2020-stronger} and sequence-to-edit tagging models \cite{omelianchuk-etal-2020-gector, lai-etal-2022-type} since introduction of the Transformer architecture \cite{vaswani2017attention}. The former generally adopt the encoder-decoder architecture \cite{lewis-etal-2020-bart}, while the latter use the encoder-only architecture \cite{devlin-etal-2019-bert}. Many state-of-the-art (SOTA) improvements are based on these two architectures, such as incorporating syntactic information into the input \cite{zhang-etal-2022-syngec} or reranking during the output process \cite{zhang-etal-2023-bidirectional, zhou-etal-2023-decoding-interventions}. Additionally, there are works that split the task into error detection and error correction phases  \cite{li-etal-2023-templategec, li-wang-2024-detection}.

Since the advent of ChatGPT \cite{brown2020gpt3}, LLMs have seen extensive application in natural language processing, particularly in text generation tasks \cite{li2024pre}. Some studies have explored the application of LLMs in GEC, such as constructing prompts for direct inference \cite{davis-etal-2024-prompting} and using parallel data for instruction tuning \cite{fan2023grammargpt, raheja-etal-2024-medit}. Research has also analyzed the performance of LLMs in the GEC task \cite{coyne2023analyzing}, including scenarios involving low-resource languages \cite{penteado2023evaluating, maeng-etal-2023-effectiveness}. The prompting strategies significantly impact performance, and there is a tendency for over-correction by LLMs \cite{loem-etal-2023-exploring, zeng-etal-2024-evaluating}. However, studies indicate that human evaluations rate the modifications made by LLMs highest \cite{coyne2023analyzing}. Our work extends the use of LLMs in multilingual GEC through an easy-to-use in-context learning approach.

\subsection{Grammatical Error Explanation}





Grammatical error explanation (GEE) refers to the elucidation of the reasons for grammatical errors in sentences and the corresponding grammatical knowledge. This requirement stems from language teaching, where providing explanations alongside error corrections is essential \cite{liang-etal-2023-chatback}. To promote the development of interpretable GEC systems, some GEC benchmarks incorporate error types, evidence words, and natural language explanations as forms of grammatical error explanation \cite{fei-etal-2023-enhancing, ye2024excgec}. \citet{kaneko-etal-2022-interpretability} utilized example-based methods to construct interpretable GEC systems. With the emergence of more LLMs with stronger linguistic capability, some works have used LLMs to generate explanations for each error in the text, and further refine the task definition and evaluation methods for GEE \cite{kaneko-okazaki-2024-controlled, song-etal-2024-gee}. Our work leverages the natural language explanations to improve the example-based GEC system.

\subsection{Demonstrations Selection}






In-context learning (ICL) is a simple and efficient method for utilizing large language models \cite{brown2020gpt3}. By using examples as demonstrations to form few-shot inference, ICL can avoid the costly optimization of large parameters. The demonstrations can be hand-crafted \cite{brown2020gpt3} or selected from labeled datasets \cite{NEURIPS2022_18abbeef}. The choice of demonstrations impacts the final performance of LLMs; therefore, some works design the selection process based on criteria such as complexity and diversity \cite{fu2022complexity, li2023finding}. Furthermore, through retrieval mechanisms, each test data can be matched with different demonstrations to enhance model performance \cite{luo2023dricl}. This process is similar to retrieval-augmented generation (RAG), where relevant information is retrieved to improve LLMs performance \cite{lewis2020rag}. Common retrieval methods involve matching inputs to inputs of the labeled data, with frequently used similarity measures including SBERT and BM25 \cite{reimers2019sentencebert, hongjin2022selective, robertson2009probabilistic}. Our work attempts to use matching between explanations as a novel retrieval method.

\section{Methods}
This section presents our design of a GEC system based on few-shot in-context learning using grammatical error explanation (GEE). First, we describe how to construct a database with GEE using labeled data in advance. Then, we explain the retrieval process based on explanations during inference, where no labels are available. Finally, we outline the few-shot template for GEC.

\begin{figure*}[ht]
    \centering
    \includegraphics[width=1.0\textwidth]{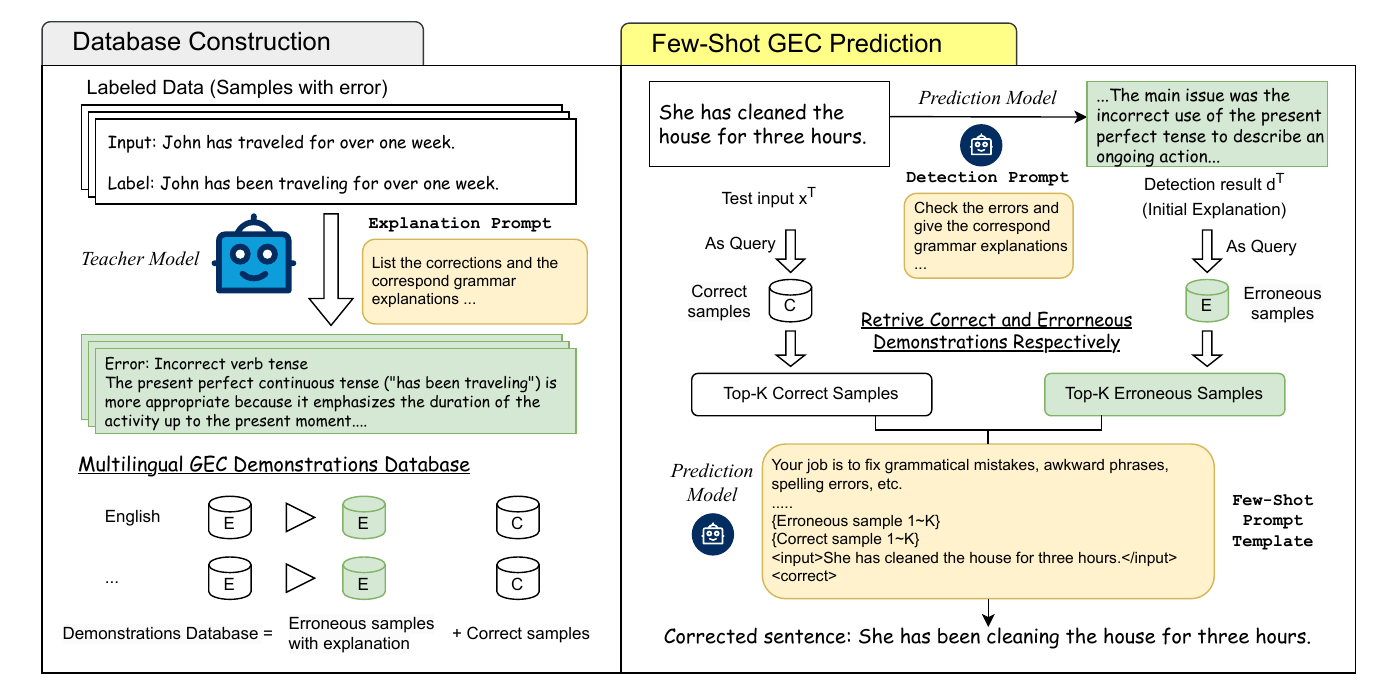}
    \caption{The proposed pipeline for few-shot GEC using the explanation-based demonstration retrieval method. As shown on the left side, we construct sample databases that include explanations. As illustrated on the right side, in the prediction stage, the erroneous samples for in-context demonstrations are retrieved using explanations.
}
    \label{fig:model}
\end{figure*}

\subsection{Explanation Database}



As the foundational and multilingual capabilities of LLMs continue to improve, LLMs can achieve strong performance on the GEE task \cite{song-etal-2024-gee}. In this paper, to ensure extensibility, we guide the LLMs to generate GEE for erroneous samples directly via prompts, as shown in Figure \ref{fig:model}. 

Building the database requires labeled datasets, which can either come from the training set of a GEC dataset or from typical grammatical correction cases manually compiled by language teachers. In this work, we choose the former as the data source, avoiding manual annotation. After applying quality filtering, we obtain a collection of samples to build the database, denoted as $\mathcal{S}$ in Equation \ref{eq:raw_dataset}.
\begin{equation}
\mathcal{S} = \{ ( x_1, y_1 ), \dots , ( x_n, y_n ) \}
\label{eq:raw_dataset}
\end{equation}

Here, $x_i$ and $y_i$ represent the \textbf{input text} (potentially erroneous) and the \textbf{corrected text} for the $i$ th sample, respectively. Next, $\mathcal{S}$ is divided into two parts: the \textbf{erroneous samples} where $x \ne y$ and the \textbf{correct samples} where $x = y$. For the former, we use an LLM, referred to as the teacher model, to generate corresponding GEE. The explanations serve as keys in the database. The database for erroneous samples is shown in Equation \ref{eq:error_database}.
\begin{equation}
(\mathcal{K},  \mathcal{V})_E  = \{ \left( e, \left(x, y \right) \right) | \forall (x, y) \in \mathcal{S}, x \ne y \}
\label{eq:error_database}
\end{equation}

Here, $\mathcal{K}$ represents the key, and $\mathcal{V}$ represents the value. The $e$ represents the \textbf{GEE} provided by the teacher model based on the pair $(x, y)$. Since GEC methods must also handle input without grammatical errors, a database for correct samples must also be constructed. As GEE cannot be generated for this part, we use the input text $x$ as the key to form the database for correct samples, as shown in Equation \ref{eq:correct_database}.
\begin{equation}
( \mathcal{K}, \mathcal{V})_C = \{ \left( x, (x, y) \right) | \forall (x, y) \in \mathcal{S}, x = y \}
\label{eq:correct_database}
\end{equation}

For each language's labeled data, we can construct two parts of the database as shown in Equations \ref{eq:error_database} and \ref{eq:correct_database}.

\subsection{Demonstrations Retrieval}

When dealing with the test data, denoted as $x^T$, the first step is to retrieve a set of samples from the database to serve as demonstrations for few-shot inference. Since the erroneous samples database uses explanations as keys, we need to rewrite the query based on the test input. Here, we use a detection prompt to guide the LLM predictor in detecting potential grammatical errors and the corresponding grammatical knowledge in the test input, which we denote as $d^T$ for an \textbf{initial explanation}. This will then serve as the query to retrieve similar samples from the erroneous samples database $(\mathcal{K}, \mathcal{V})_E$. For correct samples, we can directly use $x^T$ as the query, as shown in Equation \ref{eq:query}:
\begin{equation}
q^T_E=d^T=\mathrm{LLM}_p \left( \mathrm{prompt}_d (x^T) \right), \quad q^T_C=x^T
\label{eq:query}
\end{equation}

Here, $x^T$ represents the \textbf{test input} text that may contain grammatical errors. $q^T_E$ and $q^T_C$ are the queries used for retrieving samples from the erroneous and correct samples databases, respectively. $\mathrm{LLM}_p$ denotes the prediction model, and $\mathrm{prompt}_d$ represents the detection prompt. Once the queries are obtained, the kNN method is used to retrieve the top $k_E$ and $k_C$ samples from the two databases, as shown in Equations \ref{eq:error_demo} and \ref{eq:correct_demo}:
\begin{equation}
\mathcal{N}_E = \left\{ \left( e^{(j)} , ( x^{(j)}, y^{(j)} ) \right) \in (\mathcal{K}, \mathcal{V})_E \right\}_{j=1}^{k_E}
\label{eq:error_demo}
\end{equation}

\begin{equation}
\mathcal{N}_C = \left\{ \left( x^{(j)} , ( x^{(j)}, y^{(j)} ) \right) \in (\mathcal{K}, \mathcal{V})_C \right\}_{j=1}^{k_C}
\label{eq:correct_demo}
\end{equation}

\subsection{In-Context Learning}

In few-shot inference, the keys of the retrieved data samples are no longer used, as incorporating long explanations into the context can negatively impact the final GEC performance and significantly slow down inference. The selected demonstrations are simply the list of text pairs retrieved from the database, as shown in Equation \ref{eq:demonstration}.
\begin{equation}
D = D_E \oplus D_C
\label{eq:demonstration}
\end{equation}

Here, $D_E$ and $D_C$ are the lists of $(x, y)$ text pairs from $\mathcal{N}_E$ and $\mathcal{N}_C$, respectively. The concatenated list of samples, $D$, serves as the demonstrations, which are then combined with the few-shot prompt template for the final GEC prediction. The whole prediction process is outlined at the right half of Figure \ref{fig:model}. Given that few-shot examples can provide strong instruction-following capabilities, the prompt template is consistently written in English, with the demonstrations formatted and inserted sequentially. Further details on the prompt can be found in Appendix \ref{sec:appendix-prompt}.

\section{Experiments}
\begin{table*}[ht]
\centering
\resizebox{\textwidth}{!}{
\begin{tabular}{c|c|ccccccccccccccc}
\hline
 &  & \multicolumn{3}{c}{\textbf{English}} & \multicolumn{3}{c}{\textbf{Chinese}} & \multicolumn{3}{c}{\textbf{German}} & \multicolumn{3}{c}{\textbf{Russian}} & \multicolumn{3}{c}{\textbf{Estonian}} \\
 &  & \multicolumn{3}{c}{\textbf{CoNLL-14}} & \multicolumn{3}{c}{\textbf{NLPCC-18}} & \multicolumn{3}{c}{\textbf{Falko-Merlin}} & \multicolumn{3}{c}{\textbf{RULEC}} & \multicolumn{3}{c}{\textbf{Tartu-L1}} \\ \cline{3-17} 
\multirow{-3}{*}{\textbf{Model}} & \multirow{-3}{*}{\textbf{Method}} & \textbf{P} & \textbf{R} & \multicolumn{1}{c|}{\textbf{$\mathbf{F_{0.5}}$}} & \textbf{P} & \textbf{R} & \multicolumn{1}{c|}{\textbf{$\mathbf{F_{0.5}}$}} & \textbf{P} & \textbf{R} & \multicolumn{1}{c|}{\textbf{$\mathbf{F_{0.5}}$}} & \textbf{P} & \textbf{R} & \multicolumn{1}{c|}{\textbf{$\mathbf{F_{0.5}}$}} & \textbf{P} & \textbf{R} & \textbf{$\mathbf{F_{0.5}}$} \\ \hline
 & Random & 54.02 & 52.60 & \multicolumn{1}{c|}{53.73} & 24.87 & 16.17 & \multicolumn{1}{c|}{22.45} & 59.62 & 54.53 & \multicolumn{1}{c|}{58.53} & 36.70 & 37.45 & \multicolumn{1}{c|}{36.84} & 10.34 & 21.10 & 11.52 \\
 & Semantic & 55.21 & 51.56 & \multicolumn{1}{c|}{54.44} & 27.88 & 15.49 & \multicolumn{1}{c|}{24.04} & 60.03 & 54.15 & \multicolumn{1}{c|}{58.75} & 37.69 & 36.40 & \multicolumn{1}{c|}{37.43} & 11.04 & 22.61 & 12.30 \\
 & BM25 & 54.58 & 51.58 & \multicolumn{1}{c|}{53.95} & 24.30 & 16.41 & \multicolumn{1}{c|}{22.17} & 59.68 & 55.53 & \multicolumn{1}{c|}{58.80} & 37.77 & 36.06 & \multicolumn{1}{c|}{37.41} & - & - & - \\
\multirow{-4}{*}{\textbf{\begin{tabular}[c]{@{}c@{}}Llama3.1\\ (8B)\end{tabular}}} & \cellcolor[HTML]{EFEFEF}Explanation & \cellcolor[HTML]{EFEFEF}55.00 & \cellcolor[HTML]{EFEFEF}53.04 & \multicolumn{1}{c|}{\cellcolor[HTML]{EFEFEF}\textbf{54.60}} & \cellcolor[HTML]{EFEFEF}28.21 & \cellcolor[HTML]{EFEFEF}15.77 & \multicolumn{1}{c|}{\cellcolor[HTML]{EFEFEF}\textbf{24.36}} & \cellcolor[HTML]{EFEFEF}60.35 & \cellcolor[HTML]{EFEFEF}54.79 & \multicolumn{1}{c|}{\cellcolor[HTML]{EFEFEF}\textbf{59.15}} & \cellcolor[HTML]{EFEFEF}37.97 & \cellcolor[HTML]{EFEFEF}36.97 & \multicolumn{1}{c|}{\cellcolor[HTML]{EFEFEF}\textbf{37.76}} & \cellcolor[HTML]{EFEFEF}11.45 & \cellcolor[HTML]{EFEFEF}24.52 & \cellcolor[HTML]{EFEFEF}\textbf{12.81} \\ \hline
 & Random & 54.43 & 53.50 & \multicolumn{1}{c|}{54.24} & 27.09 & 25.34 & \multicolumn{1}{c|}{26.72} & 55.25 & 48.06 & \multicolumn{1}{c|}{53.65} & 37.79 & 40.94 & \multicolumn{1}{c|}{38.38} & 6.21 & 17.67 & 7.14 \\
 & Semantic & 55.27 & 52.65 & \multicolumn{1}{c|}{54.73} & 29.17 & 24.33 & \multicolumn{1}{c|}{28.06} & 57.81 & 48.85 & \multicolumn{1}{c|}{\textbf{55.76}} & 38.60 & 39.86 & \multicolumn{1}{c|}{38.85} & 6.41 & 18.36 & 7.37 \\
 & BM25 & 54.11 & 52.25 & \multicolumn{1}{c|}{53.73} & 28.30 & 24.95 & \multicolumn{1}{c|}{27.56} & 57.21 & 50.18 & \multicolumn{1}{c|}{55.65} & 38.03 & 40.02 & \multicolumn{1}{c|}{38.41} & - & - & - \\
\multirow{-4}{*}{\textbf{\begin{tabular}[c]{@{}c@{}}Qwen2.5\\ (7B)\end{tabular}}} & \cellcolor[HTML]{EFEFEF}Explanation & \cellcolor[HTML]{EFEFEF}55.67 & \cellcolor[HTML]{EFEFEF}51.60 & \multicolumn{1}{c|}{\cellcolor[HTML]{EFEFEF}\textbf{54.81}} & \cellcolor[HTML]{EFEFEF}29.81 & \cellcolor[HTML]{EFEFEF}23.29 & \multicolumn{1}{c|}{\cellcolor[HTML]{EFEFEF}\textbf{28.23}} & \cellcolor[HTML]{EFEFEF}57.33 & \cellcolor[HTML]{EFEFEF}47.63 & \multicolumn{1}{c|}{\cellcolor[HTML]{EFEFEF}55.08} & \cellcolor[HTML]{EFEFEF}39.16 & \cellcolor[HTML]{EFEFEF}38.48 & \multicolumn{1}{c|}{\cellcolor[HTML]{EFEFEF}\textbf{39.02}} & \cellcolor[HTML]{EFEFEF}6.54 & \cellcolor[HTML]{EFEFEF}18.21 & \cellcolor[HTML]{EFEFEF}\textbf{7.50} \\ \hline
 & Random & 63.76 & 45.68 & \multicolumn{1}{c|}{59.08} & 45.36 & 24.91 & \multicolumn{1}{c|}{38.96} & 69.18 & 46.57 & \multicolumn{1}{c|}{63.06} & 47.10 & 33.56 & \multicolumn{1}{c|}{43.59} & 19.77 & 20.26 & 19.87 \\
 & Semantic & 65.36 & 40.01 & \multicolumn{1}{c|}{58.01} & 48.37 & 21.11 & \multicolumn{1}{c|}{38.44} & 71.62 & 43.13 & \multicolumn{1}{c|}{63.27} & 45.55 & 28.28 & \multicolumn{1}{c|}{40.59} & 21.44 & 22.32 & 21.61 \\
 & BM25 & 64.30 & 38.28 & \multicolumn{1}{c|}{56.60} & 47.03 & 25.25 & \multicolumn{1}{c|}{\textbf{40.11}} & 71.58 & 42.53 & \multicolumn{1}{c|}{62.97} & 47.30 & 28.83 & \multicolumn{1}{c|}{41.93} & - & - & - \\
\multirow{-4}{*}{\textbf{Deepseek2.5}} & \cellcolor[HTML]{EFEFEF}Explanation & \cellcolor[HTML]{EFEFEF}{\color[HTML]{1F2329} 64.27} & \cellcolor[HTML]{EFEFEF}{\color[HTML]{1F2329} 45.51} & \multicolumn{1}{c|}{\cellcolor[HTML]{EFEFEF}{\color[HTML]{1F2329} \textbf{59.38}}} & \cellcolor[HTML]{EFEFEF}{\color[HTML]{1F2329} 46.92} & \cellcolor[HTML]{EFEFEF}{\color[HTML]{1F2329} 20.99} & \multicolumn{1}{c|}{\cellcolor[HTML]{EFEFEF}{\color[HTML]{1F2329} 37.62}} & \cellcolor[HTML]{EFEFEF}{\color[HTML]{1F2329} 70.73} & \cellcolor[HTML]{EFEFEF}{\color[HTML]{1F2329} 44.89} & \multicolumn{1}{c|}{\cellcolor[HTML]{EFEFEF}{\color[HTML]{1F2329} \textbf{63.43}}} & \cellcolor[HTML]{EFEFEF}{\color[HTML]{1F2329} 48.37} & \cellcolor[HTML]{EFEFEF}{\color[HTML]{1F2329} 31.52} & \multicolumn{1}{c|}{\cellcolor[HTML]{EFEFEF}{\color[HTML]{1F2329} \textbf{43.70}}} & \cellcolor[HTML]{EFEFEF}{\color[HTML]{1F2329} 22.73} & \cellcolor[HTML]{EFEFEF}{\color[HTML]{1F2329} 22.47} & \cellcolor[HTML]{EFEFEF}{\color[HTML]{1F2329} \textbf{22.68}} \\ \hline
 & Random & 57.62 & 52.38 & \multicolumn{1}{c|}{56.49} & 28.91 & 21.69 & \multicolumn{1}{c|}{27.10} & 68.22 & 50.86 & \multicolumn{1}{c|}{63.86} & 39.90 & 39.77 & \multicolumn{1}{c|}{39.87} & 19.81 & 9.40 & 16.22 \\
 & Semantic & 59.48 & 52.24 & \multicolumn{1}{c|}{57.87} & 30.72 & 21.67 & \multicolumn{1}{c|}{28.35} & 69.33 & 54.02 & \multicolumn{1}{c|}{65.61} & 41.40 & 41.96 & \multicolumn{1}{c|}{41.51} & 22.04 & 10.13 & 17.85 \\
 & BM25 & 59.21 & 51.61 & \multicolumn{1}{c|}{57.52} & 30.29 & 21.93 & \multicolumn{1}{c|}{28.14} & \cellcolor[HTML]{FFFFFF}{\color[HTML]{1F2329} 69.60} & \cellcolor[HTML]{FFFFFF}{\color[HTML]{1F2329} 51.69} & \multicolumn{1}{c|}{\cellcolor[HTML]{FFFFFF}{\color[HTML]{1F2329} 65.09}} & 40.92 & 39.69 & \multicolumn{1}{c|}{40.67} & - & - & - \\
\multirow{-4}{*}{\textbf{\begin{tabular}[c]{@{}c@{}}GPT4o\\ (mini)\end{tabular}}} & \cellcolor[HTML]{EFEFEF}Explanation & \cellcolor[HTML]{EFEFEF}60.52 & \cellcolor[HTML]{EFEFEF}52.55 & \multicolumn{1}{c|}{\cellcolor[HTML]{EFEFEF}\textbf{58.74}} & \cellcolor[HTML]{EFEFEF}31.77 & \cellcolor[HTML]{EFEFEF}21.27 & \multicolumn{1}{c|}{\cellcolor[HTML]{EFEFEF}\textbf{28.92}} & \cellcolor[HTML]{EFEFEF}69.89 & \cellcolor[HTML]{EFEFEF}52.78 & \multicolumn{1}{c|}{\cellcolor[HTML]{EFEFEF}\textbf{65.63}} & \cellcolor[HTML]{EFEFEF}42.45 & \cellcolor[HTML]{EFEFEF}41.13 & \multicolumn{1}{c|}{\cellcolor[HTML]{EFEFEF}\textbf{42.18}} & \cellcolor[HTML]{EFEFEF}25.35 & \cellcolor[HTML]{EFEFEF}10.72 & \cellcolor[HTML]{EFEFEF}\textbf{19.91} \\ \hline
\end{tabular}
}
\caption{
Results of different demonstration retrieval methods on multilingual GEC datasets. "Random" refers to random selection from database; "Semantic" and "BM25" refers to retrieval by embeddings and BM25 matching, respectively. "Explanation" is the proposed explanation-based method.
}
\label{tab:main_results}
\end{table*}




\subsection{Datasets}
\label{sec:datasets}
To evaluate the effectiveness of the proposed method, we conduct experiments on GEC datasets in multiple languages. For English, we used the W\&I+LOCNESS \cite{bryant-etal-2019-bea-19} training dataset as the database, and the CoNLL-14 \cite{ng-etal-2014-conll14} and BEA-19 datasets as the testing datasets. We also perform experiments on GEC datasets for Chinese, Russian, and German, which have a relatively high number of users. For Chinese, we utilize the HSK dataset \cite{zhang2009hsk} as the database and the NLPCC-18 dataset as the test data \cite{zhao2018nlpcc-gec}. For Russian and German, we employ the RULEC and Falko-Merlin datasets, which include both training and testing sets \cite{Rozovskaya-rulec, boyd-2018-falko_merlin}. Furthermore, we conduct experiments on the relatively low-resource Estonian language, using data from the Tartu learner corpus \footnote{\href{https://github.com/TartuNLP/estgec}{https://github.com/TartuNLP/estgec}}. In our setting, the L2 learner corpus is used as the database, while the L1 corpus serves as the test data. The database and retriever is implemented by LlamaIndex \cite{Liu_LlamaIndex_2022}.

\subsection{Models and Metrics}
\label{sec:model-metric}



We selected state-of-the-art multilingual LLMs for our experiments, including Llama-3.1-8B-Instruct by Meta \cite{dubey2024llama3} and Qwen2.5-7B-Instruct \cite{qwen2.5} developed by Tongyi. Among the closed-source models, we choose Deepseek2.5 \cite{deepseekv2} and GPT-4o-mini \cite{achiam2023gpt4-report} for its cost-effectiveness. 

In the experiments, the teacher model used to generate the GEE-based database is uniformly set to Llama-3.1-8B-Instruct. For constructing the database, we used xlm-roberta-large \cite{xlm_roberta} as the embedding model. For the prediction model in the few-shot inference stage, we conducted experiments with the four LLMs mentioned above, which serve as our primary experimental models. For the test data, we used ERRANT edit extraction results as the labels \cite{bryant-etal-2017-errant, zhang-etal-2022-mucgec}, and we evaluated the precision, recall, and $F_{0.5}$ values using the M2Scorer \cite{dahlmeier-ng-2012-m2scorer}.

The multilingual baseline methods used for comparison were: 1) \textit{Random}: randomly selecting samples from the database as demonstrations; 2) \textit{Semantic}: using the input text embedding to perform kNN retrieval \cite{khandelwal2020knnmt}; and 3) \textit{BM25}: employing BM25, a term-based retriever, for retrieval \cite{robertson2009probabilistic}. For all methods, including our proposed method, we set $k_E = 4$ erroneous samples and $k_C = 4$ correct samples as demonstrations. The final experimental results are shown in Table \ref{tab:main_results}.

\subsection{Results}

\begin{table}
\centering
\resizebox{\columnwidth}{!}{
\begin{tabular}{lllccc}
\hline
\multicolumn{1}{l|}{\multirow{2}{*}{\textbf{Method}}} & \multicolumn{1}{l|}{\multirow{2}{*}{\textbf{Backbone}}} & \multicolumn{1}{l|}{\multirow{2}{*}{\textbf{Lang}}} & \multicolumn{3}{c}{$\mathbf{F_{0.5}}$ on test set} \\ \cline{4-6} 
\multicolumn{1}{l|}{} & \multicolumn{1}{l|}{} & \multicolumn{1}{l|}{} & \textbf{EN} & \textbf{DE} & \textbf{RU} \\ \hline
\multicolumn{6}{c}{\textbf{Fine-tuned GEC Single Model}} \\ \hline
\multicolumn{1}{l|}{\citet{rothe2021clang8_gec}} & \multicolumn{1}{l|}{gT5 xxl} & \multicolumn{1}{l|}{Mono} & 65.7 & \textbf{76.0} & \textbf{51.6} \\
\multicolumn{1}{l|}{\citet{luhtaru-etal-2024-nllb-gec}} & \multicolumn{1}{l|}{NLLB} & \multicolumn{1}{l|}{Multi} &  & 73.9 &  \\
\multicolumn{1}{l|}{\citet{zhou-etal-2023-decoding-interventions}} & \multicolumn{1}{l|}{BART} & \multicolumn{1}{l|}{Mono} & \textbf{69.6} &  &  \\ \hline
\multicolumn{6}{c}{\textbf{Inference of LLMs}} \\ \hline
\multicolumn{1}{l|}{\citet{davis-etal-2024-prompting}} & \multicolumn{1}{l|}{GPT-3.5-Turbo} & \multicolumn{1}{l|}{-} & 57.2 &  &  \\
\multicolumn{1}{l|}{\citet{tang-etal-2024-ungrammatical}} & \multicolumn{1}{l|}{GPT-3.5-Turbo} & \multicolumn{1}{l|}{-} & 58.8 &  &  \\
\multicolumn{1}{l|}{Ours} & \multicolumn{1}{l|}{Deepseek2.5} & \multicolumn{1}{l|}{-} & \textbf{59.4} & \textbf{63.4} & \textbf{43.7} \\ \hline
\end{tabular}
}
\caption{
Some reported metrics for state-of-the-art (SOTA) multilingual GEC methods based on fine-tuning or direct inference. "EN", "DE", and "RU" denote the CoNLL-14, Falko-Merlin, and RULEC datasets, respectively. For the fine-tuning methods, some models are trained using multilingual mixed data, indicated as "Multi" in the "Lang" column, while other methods fine-tune models separately for each language, marked as "Mono".
}
\label{tab:sota}
\end{table}

\noindent \textbf{Performance of the proposed method} \quad From Table \ref{tab:main_results}, we can see that when using $F_{0.5}$ as the evaluation metric, our proposed method generally outperforms other reproduced methods under the same experimental setup. Overall, the proposed method significantly outperforms random selection. This is particularly evident with GPT4o-mini, where the $F_{0.5}$ score of the proposed method exceeds that of \textit{Random} by approximately 2 points. Among the baseline methods, the \textit{Semantic} method performs best with open-source LLMs, averaging about 1 point higher than the \textit{Random} method. However, this is not the case for the Deepseek2.5, possibly due to randomness introduced by the temperature setting. As the representative models, Llama3.1 (open-source) and GPT4o-mini (closed-source), the proposed method's average $F_{0.5}$ value exceeds that of the \textit{Semantic} method by 0.36 and 0.84, respectively. Given that the explanation texts in erroneous samples differ entirely from the input text, this result shows that using explanations as a tool for in-context demonstration retrieval is indeed a feasible approach, which is slightly more effective than directly matching based on input text embeddings. Several examples demonstrating improvements have been included in the Appendix \ref{sec:appendix-cases}.

\noindent \textbf{In-context learning performance for multilingual GEC} \quad From Table \ref{tab:main_results}, we can observe that both open-source and closed-source LLMs demonstrate certain multilingual GEC capabilities. However, their performances vary significantly across different languages, possibly due to the different capability among languages and the different levels of difficulty of the datasets. for instance, the L1 learner dataset for Estonian is more challenging, resulting in lower performance metrics across all models. Overall, closed-source models tend to outperform open-source models with around 8B parameters, which is reasonable given that closed-source models often have larger architectures and pre-training scales. Additionally, it is noteworthy that, in many cases, regardless of the method used to select demonstrations, the few-shot performance of large models does not differ significantly. For example, when using Llama3.1 on the Falko-Merlin dataset, the $F_{0.5}$ scores differ by only 0.6 across 4 demonstration retrieval strategies. These results indicate that the GEC capabilities of large models largely depend on their inherent abilities, while the selection of examples has some influence around the inherent few-shot performance level.

\noindent \textbf{Comparison with SOTA} \quad We compared the performance of the LLMs few-shot inference with several state-of-the-art (SOTA) multilingual GEC methods, as shown in Table \ref{tab:sota}. Most SOTA methods are all fine-tuned on labeled datasets, with some methods tuning a generative model for each language individually. Only results using the same test sets are listed in the table for comparison. We can see that current few-shot ICL lags behind supervised finetuning (SFT) by about 10 points in terms of $F_{0.5}$. Considering that the GEC performance of LLMs is actually underestimated \cite{coyne2023analyzing}, this gap may be smaller. As the capabilities of LLMs continue to improve, we can expect that in the future, ICL could replace SFT as the most efficient method for multilingual GEC.

\begin{table*}[ht]
\centering
\resizebox{\textwidth}{!}{
\begin{tabular}{c|c|ccccccccccccccc}
\hline
\multirow{2}{*}{\textbf{Shot}} & \multirow{2}{*}{\textbf{Explanation}} & \multicolumn{3}{c}{\textbf{English}} & \multicolumn{3}{c}{\textbf{Chinese}} & \multicolumn{3}{c}{\textbf{German}} & \multicolumn{3}{c}{\textbf{Russian}} & \multicolumn{3}{c}{\textbf{Estonian}} \\ \cline{3-17} 
 &  & \textbf{P} & \textbf{R} & \multicolumn{1}{c|}{\textbf{$\mathbf{F_{0.5}}$}} & \textbf{P} & \textbf{R} & \multicolumn{1}{c|}{\textbf{$\mathbf{F_{0.5}}$}} & \textbf{P} & \textbf{R} & \multicolumn{1}{c|}{\textbf{$\mathbf{F_{0.5}}$}} & \textbf{P} & \textbf{R} & \multicolumn{1}{c|}{\textbf{$\mathbf{F_{0.5}}$}} & \textbf{P} & \textbf{R} & \textbf{$\mathbf{F_{0.5}}$} \\ \hline
\multirow{2}{*}{\textbf{0-Shot}} & - & 51.16 & 53.95 & \multicolumn{1}{c|}{51.70} & 17.48 & 19.91 & \multicolumn{1}{c|}{17.92} & 54.75 & 55.69 & \multicolumn{1}{c|}{54.93} & 28.96 & 41.36 & \multicolumn{1}{c|}{30.81} & 6.87 & 20.41 & 7.92 \\
 & \textbf{Pre} & 41.16 & 47.22 & \multicolumn{1}{c|}{42.24} & 9.84 & 15.17 & \multicolumn{1}{c|}{10.58} & 47.16 & 48.83 & \multicolumn{1}{c|}{47.48} & 19.38 & 33.54 & \multicolumn{1}{c|}{21.17} & 4.83 & 18.06 & 5.66 \\ \hline
\multirow{3}{*}{\textbf{8-Shot}} & - & 55.00 & 53.04 & \multicolumn{1}{c|}{\textbf{54.60}} & 28.21 & 15.77 & \multicolumn{1}{c|}{\textbf{24.36}} & 60.35 & 54.79 & \multicolumn{1}{c|}{\textbf{59.15}} & 37.97 & 36.97 & \multicolumn{1}{c|}{37.76} & 11.45 & 24.52 & \textbf{12.81} \\
 & \textbf{Pre} & 40.62 & 44.11 & \multicolumn{1}{c|}{41.27} & 10.35 & 14.57 & \multicolumn{1}{c|}{10.99} & 47.60 & 47.48 & \multicolumn{1}{c|}{47.57} & 19.91 & 31.86 & \multicolumn{1}{c|}{21.53} & 5.09 & 18.36 & 5.95 \\
 & \textbf{Post} & 54.42 & 51.46 & \multicolumn{1}{c|}{53.80} & 28.06 & 14.84 & \multicolumn{1}{c|}{23.81} & 59.84 & 53.38 & \multicolumn{1}{c|}{58.43} & 38.74 & 36.66 & \multicolumn{1}{c|}{\textbf{38.31}} & 10.87 & 23.05 & 12.15 \\ \hline
\end{tabular}
}
\caption{
Results of incorporating explanations into context. The test datasets are the same as those in Table \ref{tab:main_results}. The explanation can be placed before the corrected text (pre) or after the corrected text (post). The performance of zero-shot (Row 1 of metrics) and the proposed method (Row 3) is provided for comparison. 
}
\label{tab:cot}
\end{table*}

\section{Discussion}
\subsection{The Extension to Other Datasets}

\begin{table}
\centering

\resizebox{\columnwidth}{!}{
\begin{tabular}{c|cccccc}
\hline
 & \multicolumn{3}{c}{\textbf{Llama3.1}(8B)} & \multicolumn{3}{c}{\textbf{Qwen2.5}(7B)} \\ \cline{2-7} 
\multirow{-2}{*}{\textbf{Method}} & \textbf{P} & \textbf{R} & \textbf{$\mathbf{F_{0.5}}$} & \textbf{P} & \textbf{R} & \textbf{$\mathbf{F_{0.5}}$} \\ \hline
Random & 44.2 & 63.4 & 47.1 & 44.8 & 63.6 & 47.7 \\
Semantic & 45.5 & 62.8 & 48.1 & 45.5 & 63.4 & 48.2 \\
BM25 & 44.2 & 63.0 & 47.0 & 44.7 & 63.9 & 47.6 \\
\rowcolor[HTML]{EFEFEF} 
Explanation & {\color[HTML]{1F2329} 45.2} & {\color[HTML]{1F2329} 63.3} & {\color[HTML]{1F2329} \textbf{48.1}} & \cellcolor[HTML]{EFEFEF}{\color[HTML]{1F2329} 47.2} & \cellcolor[HTML]{EFEFEF}{\color[HTML]{1F2329} 62.3} & \cellcolor[HTML]{EFEFEF}{\color[HTML]{1F2329} \textbf{49.6}} \\ \hline
\end{tabular}
}
\caption{
Results of different demonstration retrieval methods on BEA-19 test set (English).
}
\label{tab:transfer}
\end{table}

To verify that the constructed database is effective across multiple test sets of the same language, we conducted experiments on another commonly used English GEC test set, BEA-19 \cite{bryant-etal-2019-bea-19}, using Llama3.1 and Qwen2.5. As shown in Table \ref{tab:transfer}, the methods exhibit similar trend to those observed in the main experiments. On the Qwen2.5 model, the advantage of the proposed method is more pronounced. These results demonstrate that the explanation-based erroneous samples database we constructed can be transferred to different test scenarios to a certain extent while maintaining the advantage in performance.

\subsection{The Performance of In-context Explanations}
\label{sec:in-context-explanation}
A key point in using explanations for demonstration retrieval is how to generate a query for retrieval when the corrected text is unknown. We employ a detection prompt to perform grammar checking on the test input, obtaining an initial explanation to serve as the query. This raises the question: \textit{why not directly use this initial explanation as the chain of thought (CoT) in the reasoning process and let the model derive the final GEC answer based on it?} We experimented with this CoT approach, and as shown in Table \ref{tab:cot}, the first two rows present the results of the zero-shot and CoT experiments with the Llama3.1 model. The significant drop in the $F_{0.5}$ score indicates that this CoT method is completely ineffective for GEC. Possible reasons include: 1) Without labeled data, the quality of the initial explanation is poor, making it only useful as a retrieval tool but misleading when used as an intermediate reasoning step, especially given the minimal difference between input and output in GEC tasks; 2) The intermediate reasoning step may interfere with the model’s ability to follow instructions for the final output.

Given the poor performance of CoT in the zero-shot scenario, \textit{What about adding explanations in the context of the few-shot inference?} We conduct experiments on this as well. There are two possible placements for the explanation text: one is before the answer, following an input-explanation-output pattern, and the other is after the answer, following an input-output-explanation pattern. In the experiments, the same format is applied across all examples and test data, with explanations set to "No error in text" for error-free samples. Notably, in the latter pattern, the predicted sample directly outputs the corrected text without using the initial explanation as an intermediate reasoning step as in the former case. The results in rows 4 and 5 of Table \ref{tab:cot} show that adding explanation text generally degrades GEC performance. This may be because the lengthy explanation takes up a large portion of the context, which negatively impacts the LLMs' performance on the GEC task itself. Moreover, in row 4, we observe a trend similar to that in row 2, further validating that using the initial explanation as an intermediate reasoning step weakens the results. The proposed method of using it for demonstration retrieval, rather than reasoning, is one of the correct ways to utilize it.

\subsection{The Balance of Erroneous and Correct Demonstrations}

\begin{figure*}[ht]
    \centering
    \includegraphics[width=1.0\textwidth]{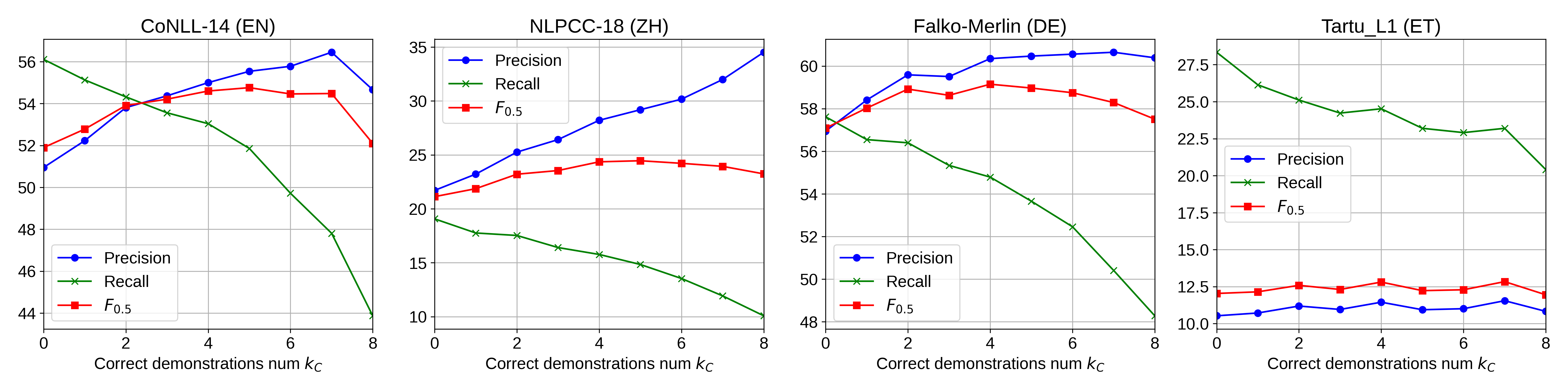}
    \caption{GEC metrics on 4 datasets as the number of the correct samples $k_C$ varies, with a total of $k_E + k_C = 8$ in-context demonstrations consisting of both erroneous and correct samples.}
    \label{fig:trade-off}
\end{figure*}

As described in Section \ref{sec:model-metric}, the number of correct and erroneous samples in the demonstrations is set to 4 each. The ratio between these two types of samples can affect precision and recall, thereby influencing the final GEC performance measured by $F_{0.5}$. By fixing the total number of samples at 8 and varying the number of correct samples from 0 to 8, we can create a series of experiments to observe how GEC performance changes with different ratios of correct to erroneous samples. As shown in Figure \ref{fig:trade-off}, as the number of correct samples increases and the number of erroneous samples decreases, there is a general trend of increasing precision and decreasing recall. The $F_{0.5}$ score achieves its maximum value when the number of correct samples is 4 or 5. Therefore, for ICL of multilingual GEC, it is reasonable to use an equal number of correct and erroneous samples, which is also our setting.

\subsection{The Guidance of Explanation Generation}
\label{sec:exp_prompt_influence}

\begin{table}
\centering
\small
\resizebox{\columnwidth}{!}{
\begin{tabular}{l|ccc|c}
\hline
\multirow{2}{*}{\textbf{Dataset}} & \multicolumn{3}{c|}{\textbf{Explanation (Using Edits)}} & \textbf{Explanation} \\ \cline{2-5} 
 & \textbf{P} & \textbf{R} & \textbf{$\mathbf{F_{0.5}}$} & \textbf{$\mathbf{F_{0.5}}$} \\ \hline
\textbf{English} & 55.30 & 54.09 & \textbf{55.05} & 54.60 \\
\textbf{Chinese} & 28.23 & 15.22 & 24.28 & \textbf{24.36} \\
\textbf{German} & 60.44 & 54.85 & \textbf{59.23} & 59.15 \\
\textbf{Russian} & 37.85 & 37.54 & \textbf{37.79} & 37.76 \\
\textbf{Estonian} & 10.95 & 26.14 & 12.39 & \textbf{12.81} \\ \hline
\end{tabular}
}
\caption{
The performance of the system using the explanation generated by the prompt template with edits extracted by alignment, compared with the direct explanation of the proposed method. The test datasets are the same as those in Table \ref{tab:main_results}.
}
\label{tab:exp-prompt}
\end{table}

In previous works on generating grammatical error explanations (GEE), the approach typically involved extracting edit operations by aligning input text with corrected text \cite{kaneko-okazaki-2024-controlled, song-etal-2024-gee}. In our proposed method, we directly allow the teacher model to generate GEE based on the original text pairs. Additionally, we also crafted a version that utilizes extracted edits in the explanation prompt and conduct the experiments on LLama3.1. Two prompts are both shown in Appendix \ref{sec:appendix-prompt}. The results are shown in the first three columns of Table \ref{tab:exp-prompt}, with the last column representing the $F_{0.5}$ performance of the proposed method in the main experiment. It is evident that both types of explanations have their strengths and weaknesses. Ultimately, we opted for the GEE generated directly based on the original text pairs, as correcting a grammatical error may involve multiple edit operations. Not specifying the edit operations in advance allows LLMs greater freedom in generating explanations.



\section{Conclusion}

In conclusion, this paper presents a novel approach for retrieving demonstrations based on grammatical error explanations (GEE) to enhance in-context learning (ICL) for the multilingual grammatical error correction (GEC) task. Our method addresses the challenge of selecting effective in-context demonstrations by matching GEE within a constructed demonstration database, improving the few-shot performance of LLMs in multilingual GEC without requiring additional training. 

Future work will explore: 1) methods for indexing correct samples with suitable explanations, which was not fully addressed in this work; 2) refining the retrieval method. In this study, only text embeddings were used as representations for explanations; and 3) Further refinement of GEE. A more systematic form of GEE and its evaluation methodologies would facilitate the evolution of retrieval systems based on this framework. Additionally, it would also promote the application of GEE within the domain of GEC.

We believe that explanation-based demonstration retrieval has broader implications for ICL across all natural language processing downstream tasks. For GEC, we consider few-shot inference to be the most effective approach for building an explainable multilingual GEC system using LLMs. As LLMs' multilingual capabilities continue to improve, we expect that few-shot ICL will become an efficient and cost-effective solution for GEC.

\section*{Limitations}
Firstly, due to limitations in computational resources, we only used datasets from 5 languages and conducted experiments on 2 open-source LLMs and 1 closed-source LLM. To further validate the effectiveness of the proposed method, experiments should be conducted on a wider variety of LLMs with different sizes and on more datasets across more languages. Secondly, in this paper, the teacher model used to generate the explanation database is the open-source Llama3.1 with 8B parameters. Since generating the database requires high-quality explanations, using a closed-source LLM with stronger base capabilities could be a better choice and may further improve the performance of the proposed method. Although the database can be reused once it is built, its large scale makes the cost of using closed-source LLMs to build it relatively high. Additionally, we only compared the proposed method with 3 baseline methods. There are indeed many ICL demonstration retrieval methods, but many are implemented for specific tasks or specific languages, making it challenging to reproduce and apply them to multilingual GEC. For example, BM25 retrieval based on LlamaIndex encountered issues with the Estonian dataset, leaving this part of the experiment incomplete.

The current research on GEE also limits the performance of the method proposed in this paper. Currently, there is no systematic and automated method for evaluating the quality of GEE, and there are no unified standards regarding the contents that should be included within GEE. This paper can only demonstrate that the quality of explanations generated by LLMs is sufficient to support improvements in the performance of the few-shot GEC. We hope that future work will involve the elaboration for GEE.

\section*{Ethics Statement}
The datasets and models used in this work are publicly available and have been employed exclusively for research purposes. The datasets do not contain any personally identifiable information or offensive content. The large language models (LLMs) utilized in our experiments are consistent with their permitted use in academic research. The code for our proposed method will be made publicly available for academic research in grammatical error correction, adhering to the access conditions of the LLMs used in this study.

LLMs are almost unlikely to generate harmful content under our few-shot settings, as the input remains largely unchanged in the grammatical error correction (GEC) task. However, there is still potential risk of hallucinations, particularly when certain words involving facts are modified by the models.

Additionally, GPT-4o from the ChatGPT platform was used as an AI assistant for refining the paper writing.

\section*{Acknowledgments}
This work was supported by National Science and Technology Major Project (No. 2022ZD0116308)  and National Natural Science Foundation of China (62036001) . The corresponding author is Houfeng Wang.

\bibliography{custom}

\appendix

\section{Dataset Statistics}

\begin{table*}
\centering
\small
\begin{tabular}{l|lll|ll}
\hline
 & \multicolumn{3}{c|}{\textbf{Database}} & \multicolumn{2}{c}{\textbf{Test Dataset}} \\ \hline
\textbf{Language} & \textbf{Origin} & \textbf{\#Erroneous} & \textbf{\#Correct} & \textbf{Origin} & \textbf{\#Total} \\ \hline
\textbf{English} & W\&I+LOCNESS & 20185 & 6839 & CoNLL-14 & 1312 \\
\textbf{Chinese} & HSK & 25000* & 25000* & NLPCC-18 & 2000 \\
\textbf{German} & Falko-Merlin & 11801 & 1916 & Falko-Merlin & 2337 \\
\textbf{Russian} & RULEC & 961 & 913 & RULEC & 5000 \\
\textbf{Estonian} & Tartu-L2-Corpus & 7156 & 2** & Tartu-L1-Corpus & 1453 \\ \hline
\end{tabular}
\caption{
The GEC data quantity used. For the database, \#Erroneous represents the number of erroneous samples, and \#Correct represents the number of correct samples. For the test data, \#Total indicates the total number of samples. *For the HSK dataset, due to its large size, we randomly selected 25,000 erroneous and 25,000 correct samples for database construction. **The Tartu-L2-Corpus contains only 4 correct samples, and after filtering, 2 samples remained.
}
\label{tab:statistics}
\end{table*}

As described in Section \ref{sec:datasets}, our dataset usage consists of two parts: the labeled data used to build the database and the test data used to evaluate the proposed method. The labeled data samples used to construct the database are initially filtered by length, with a minimum of 10 and a maximum of 60 tokens to ensure quality. Additionally, we limited the number of total samples to 25,000. The filtering process also helped to reduce the cost of constructing the explanation database. The statistics of the datasets used in this paper are shown in Table \ref{tab:statistics}.

\section{Experimental settings}
\subsection{Model Settings}

We conducted experiments using open-source LLMs available on the Huggingface community, including \href{https://huggingface.co/meta-llama/Llama-3.1-8B-Instruct}{Llama-3.1-8B-Instruct} and \href{https://huggingface.co/Qwen/Qwen2.5-7B-Instruct}{Qwen2.5-7B-Instruct}. During generation, we set the temperature to 0 and avoided any random sampling strategies to eliminate output randomness for all open-source LLMs. For closed-source LLMs, we utilized their official APIs and applied their default settings to get better performance, which introduced some degree of randomness. As a baseline for comparison, the method of "Random" selecting samples for open-source LLMs is evaluated by running it with 3 different random seeds and reporting the average result, while the other method will get a fixed output so only one-round inference is performed. Due to the cost of usage, only the single-round results of API-based closed LLMs are reported.

\subsection{Prompt Settings}
\label{sec:appendix-prompt}
\begin{table*}[ht]
\centering
\resizebox{\textwidth}{!}{
\begin{tabular}{l|l|l}
\hline
\textbf{Prompt} & \textbf{Usage} & \textbf{Prompt Content} \\ \hline
\textbf{Explanation} & \textbf{Teacher Model} & \begin{tabular}[c]{@{}l@{}}You are tasked with performing a comprehensive grammatical error analysis on the following text, which may contain errors in various languages. \\ Your job is to identify any grammatical, syntactic, punctuation, or spelling errors in the text. For each detected error, follow these steps:\\ \\ Identify the Error: List each error separately.\\ Correction: Suggest the appropriate correction for each identified error.\\ Explanation: Provide a brief explanation for why the correction is necessary. This should include references to specific grammar rules, conventions, \\ or language-specific nuances (e.g., verb conjugation, article-noun agreement, preposition use, punctuation rules).\\ For spelling errors, offer an explanation if the mistake could arise from common language-specific confusions (e.g., homophones, loanwords).\\ For punctuation issues, explain the relevant punctuation rules (e.g., comma placement in subordinate clauses, quotation marks, etc.).\\ For syntax or word order issues, explain how sentence structure works in the language and why the original sentence does not follow the norm.\\ Minimal Impact on Meaning: Ensure that the corrections you propose do not alter the original meaning of the sentence. The goal is to preserve the \\ intent of the writer while correcting errors.\\ When explaining each error, keep in mind that the explanations should be clear and concise but still detailed enough to be educational. Whenever \\ possible, reference grammatical terms (e.g., agreement, tense, case, gender, aspect) relevant to the error.\\ \\ Important Notes:\\ \\ If the text is multilingual, address each language's grammar rules separately.\\ Your explanations should cater to a general audience, meaning that while your responses can be technical, they should still be easily understood by \\ someone with a basic understanding of grammar.\\ Now, perform this process for the following text:\\ \\ {[}The given text{]}:\\ \{text\}\\ \\ And we will give you the corrected version of the given text below. Your analysis of grammatical errors should lead to the given text being corrected \\ to this specific version.\\ {[}The corrected version{]}:\\ \{label\}\\ \\ {[}Corrections made and the brief reasons for the errors{]}:\end{tabular} \\ \hline
\textbf{\begin{tabular}[c]{@{}l@{}}Explanation\\ (Using Edits)\end{tabular}} & \textbf{Teacher Model} & \begin{tabular}[c]{@{}l@{}}You, a language expert, can briefly explain how to judge a sentence is grammatically correct and why some corrections are essential.\\ For the given text:\\ \{text\}\\ Corrected text:\\ \{label\}\\ \\ \{Extracted edits here, like: insert "xxx" between "yyy" and "zzz"; replace "xxx" with "yyy".\}\\ Please explain briefly why you made these corrections.\\ Your Explanation:\end{tabular} \\ \hline
\textbf{\begin{tabular}[c]{@{}l@{}}Detection\\ (Detailed)\end{tabular}} & \textbf{\begin{tabular}[c]{@{}l@{}}Prediction Model\\ Llama3.1\\ Deepseek2.5\\ GPT4o-mini\end{tabular}} & \begin{tabular}[c]{@{}l@{}}You are tasked with performing a comprehensive grammatical error analysis on the following text, which may contain errors in various languages. \\ Your job is to identify any grammatical, syntactic, punctuation, or spelling errors in the text. For each detected error, follow these steps:\\ \\ Identify the Error: List each error separately.\\ Correction: Suggest the appropriate correction for each identified error.\\ Explanation: Provide a brief explanation for why the correction is necessary. This should include references to specific grammar rules, conventions, \\ or language-specific nuances (e.g., verb conjugation, article-noun agreement, preposition use, punctuation rules).\\ For spelling errors, offer an explanation if the mistake could arise from common language-specific confusions (e.g., homophones, loanwords).\\ For punctuation issues, explain the relevant punctuation rules (e.g., comma placement in subordinate clauses, quotation marks, etc.).\\ For syntax or word order issues, explain how sentence structure works in the language and why the original sentence does not follow the norm.\\ Minimal Impact on Meaning: Ensure that the corrections you propose do not alter the original meaning of the sentence. The goal is to preserve the \\ intent of the writer while correcting errors.\\ When explaining each error, keep in mind that the explanations should be clear and concise but still detailed enough to be educational. Whenever \\ possible, reference grammatical terms (e.g., agreement, tense, case, gender, aspect) relevant to the error.\\ \\ Important Notes:\\ \\ If the text is multilingual, address each language's grammar rules separately.\\ Your explanations should cater to a general audience, meaning that while your responses can be technical, they should still be easily understood by \\ someone with a basic understanding of grammar.\\ Now, perform this process for the following text:\\ \\ {[}The given text{]}:\\ \{source\}\\ \\ {[}Corrections made and the brief reasons for the errors{]}:\end{tabular} \\ \hline
\textbf{\begin{tabular}[c]{@{}l@{}}Detection\\ (Short)\end{tabular}} & \textbf{\begin{tabular}[c]{@{}l@{}}Prediction Model\\ Qwen2.5\end{tabular}} & \begin{tabular}[c]{@{}l@{}}Your task is to detect grammatical errors in the given text and provide corrections along with explanations based on the relevant grammar rules. \\ For each error found, specify the type of error (e.g., subject-verb agreement, tense inconsistency) and explain why it is incorrect. \\ Then provide the correct version of the sentence and briefly explain the grammar rule that applies.\\ \\ Please follow this structure for your response:\\ \\ {[}The given text{]}:\\ \{source\}\\ \\ {[}Corrections made and the brief reasons for the errors{]}:\end{tabular} \\ \hline
\textbf{Few-Shot} & \textbf{\begin{tabular}[c]{@{}l@{}}Prediction Model\\ All\end{tabular}} & \begin{tabular}[c]{@{}l@{}}You are an language expert who is responsible for grammatical, lexical and orthographic error corrections given an input sentence. Your job is to \\ fix grammatical mistakes, awkward phrases, spelling errors, etc. following standard written usage conventions, but your corrections must be \\ conservative. Please keep the original sentence (words, phrases, and structure) as much as possible. The ultimate goal of this task is to make the \\ given sentence sound natural to native speakers without making unnecessary changes. Corrections are not required when the sentence is already \\ grammatical and sounds natural.\\ There is an erroneous sentence between `\textless{}erroneous sentence\textgreater{}` and `\textless{}/erroneous sentence\textgreater{}`. Then grammatical errors in the erroneous sentence\\ will be corrected. The corrected version will be between `\textless{}corrected sentence\textgreater{}` and `\textless{}/corrected sentence\textgreater{}`.\\ \textless{}erroneous sentence\textgreater{}\{text\}\textless{}/erroneous sentence\textgreater\\ \textless{}corrected sentence\textgreater{}\{label\}\textless{}/corrected sentence\textgreater\\ ...\\ \textless{}erroneous sentence\textgreater{}\{text\}\textless{}/erroneous sentence\textgreater\\ \textless{}corrected sentence\textgreater{}\{label\}\textless{}/corrected sentence\textgreater\\ \textless{}erroneous sentence\textgreater{}\{source\}\textless{}/erroneous sentence\textgreater\\ \textless{}corrected sentence\textgreater{}\end{tabular} \\ \hline
\end{tabular}
}
\caption{
The prompts for the proposed method. \{text\} and \{label\} means the input text and correct sentence (label) for labeled GEC data. \{source\} represents the test input text.
}
\label{tab:prompts}
\end{table*}

Three types of prompts are used in our experiments, as illustrated in Figure \ref{fig:model}. During the database construction phase, an explanation prompt is employed to generate explanations. In the prediction phase, a detection prompt is used to provide an initial explanation when no ground truth correction is available. The few-shot prompt template is used for the final few-shot inference incorporating demonstrations, and it is applied both in the comparative methods and the proposed method. Table \ref{tab:prompts} displays the actual prompts used in our experiments. For the detection prompt, we leveraged GPT-4o\footnote{\href{https://chatgpt.com}{https://chatgpt.com}} to generate both detailed and short version. As for the explanation prompt, one approach was to directly combine the detection prompt with the corrected sentence, which served as our primary method. In Section \ref{sec:exp_prompt_influence}, we also explored an edit-guided explanation prompt. The few-shot prompt was formulated based on prior work \cite{tang-etal-2024-ungrammatical, davis-etal-2024-prompting}.

\section{Case Study}
\label{sec:appendix-cases}
\begin{figure*}[ht]
    \centering
    \includegraphics[width=1.0\textwidth]{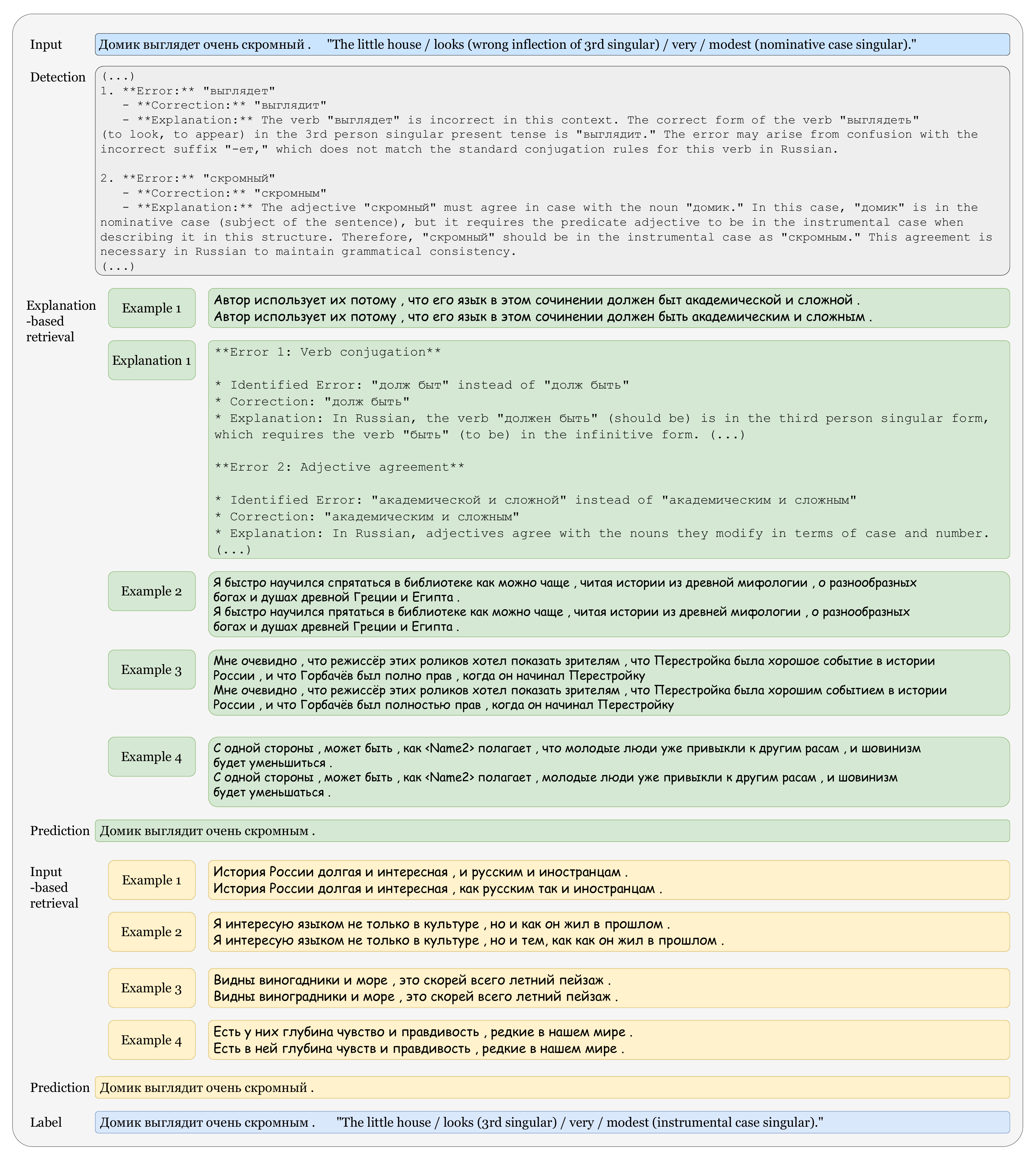}
    \caption{An example of the retrieval and generation result from the Russian GEC.}
    \label{fig:case-russian}
\end{figure*}

\begin{figure*}[ht]
    \centering
    \includegraphics[width=1.0\textwidth]{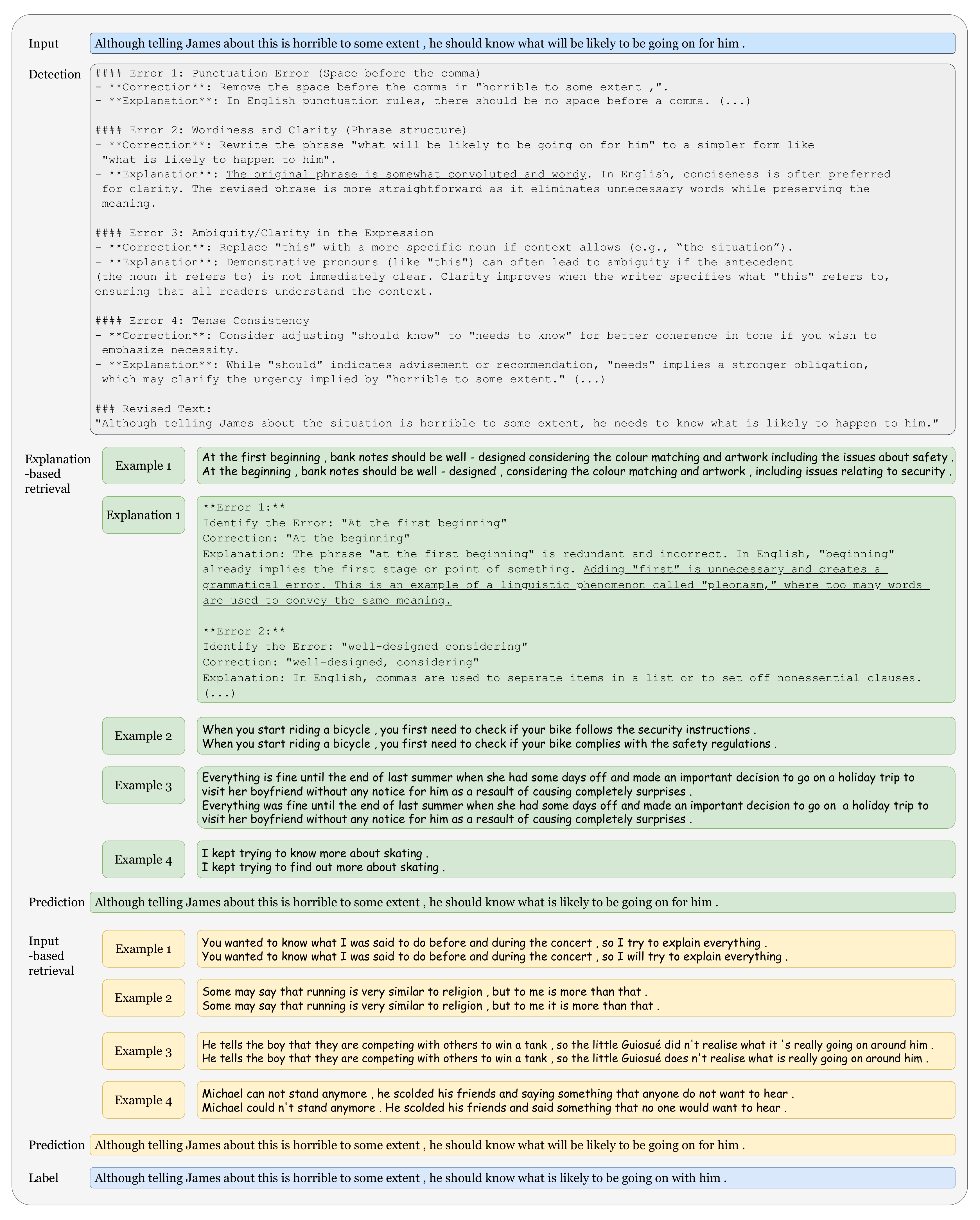}
    \caption{An example of the retrieval and generation result from the English GEC.}
    \label{fig:case-english}
\end{figure*}

In this section, two cases are presented in which explanation-based retrieval exerts a significant influence in improving the prediction. Due to space limitations, for each input, only the detection result and the explanation of the first retrieved sample will be shown. In Figures \ref{fig:case-russian} and \ref{fig:case-english}, we present the detection results generated from the input sentences. Additionally, we illustrate the retrieval results of erroneous samples using explanation-based retrieval and input-based ("Semantic" method) retrieval, along with the few-shot prediction results when using the retrieved examples as context.

As shown in Figure \ref{fig:case-russian}, the input Russian sentence contains errors in verb conjugation and adjective declension. The detection result produced by the LLMs (referred to as the initial explanation in the proposed method) accurately identified these errors. The first sample retrieved based on this explanation not only includes both types of errors but also shares similar structure around the erroneous part with the input sentence. In contrast, samples retrieved based on the input alone contained different types of errors, which led to unsuccessful correction of the adjective declension in particular. Using the samples retrieved based on the explanation, all errors were correctly corrected.

In practice, the results of detection and explanation-based retrieval are not always perfect; however, they can still improve model performance in certain scenarios. As shown in Figure \ref{fig:case-english}, which illustrates a case in English GEC, the detection result includes multiple errors identified by LLMs. Among these, the "wordy" error is accurately identified and required correction. The first sample retrieved using this initial explanation indeed contained a similar error, with the segment highlighted by an underline to indicate the similarity of GEE. In contrast, samples retrieved based on the input do not contain similar errors, resulting in no modifications being made in the final few-shot prediction. Utilizing the sample retrieved based on the explanation, the few-shot prediction successfully corrected the "wordy" error of "will be likely" but failed to correct the preposition error at the end of the sentence. Interestingly, despite multiple errors being detected, the LLMs did not exhibit a tendency towards overcorrection under the influence of in-context learning. This observation partly explains why incorporating explanations into the context, as discussed in Section \ref{sec:in-context-explanation}, will lead to a degradation in performance compared to simply placing samples in context.

\end{document}